\documentclass{article} 
\usepackage{iclr2016_conference,times}
\usepackage{hyperref}
\usepackage{url}
 \usepackage{graphicx}
 \usepackage{enumerate}
 \usepackage{amsmath}
 \usepackage{multirow}

\title{Hierarchical Latent Semantic Mapping for Automated Topic Generation}

\author{Guorui Zhou, Guang Chen \\
School of Information and Communication Engineering\\
Beijing University of Posts and Telecommunications\\
Beijing, China \\
\texttt{\{zhouguorui, chenguang\}@bupt.edu.cn} \\
}

\begin{document}

\maketitle

\begin{abstract}
Much of information sits in an unprecedented amount of text data. Managing allocation of these large scale text data is an important problem for many areas. Topic modeling performs well in this problem. The traditional generative models (PLSA,LDA) are the state-of-the-art approaches in topic modeling and most recent research on topic generation has been focusing on improving or extending these models. However, results of traditional generative models are sensitive to the number of topics $K$, which must be specified manually and determines the $rank$ of solution space for topic generation. The problem of generating topics from corpus resembles community detection in networks. Many effective algorithms can automatically detect communities from networks without a manually specified number of the communities. Inspired by these algorithms, in this paper, we propose a novel method named Hierarchical Latent Semantic Mapping (HLSM), which $automatically$ generates topics from corpus. HLSM calculates the association between each pair of words in the latent topic space, then constructs a unipartite network of words with this association and hierarchically generates topics from this network. We apply HLSM to several document collections and the experimental comparisons against several state-of-the-art approaches demonstrate the promising performance.
\end{abstract}

\section{Introduction}
Managing large allocation of documents has become a popular challenge in many fields. Topic modeling, which assigns topics to documents, offers a promising solution for this challenge.

  Topic models generate topics from a set of documents and assign topics to these documents. Based on these topics we can solve problems on cross-domain text classification \citet{Ctcusba}, understanding text clustering \citet{TopicClust}, text recommendation, and other related text data applications. There has been an exceptional amount of research on topic-model algorithms. PLSA and LDA are highly modular and can therefore be easily extended. Since LDA's introduction, there is much research based on it. The Correlated Topic Model  Advances \citet{Blei:2006:DTM:1143844.1143859} follows this approach, inducing a correlation structure between topics by using the logistic normal distribution instead of the Dirichlet. Another extension is the hierarchical LDA \citet{Blei:2010:NCR:1667053.1667056}, where topics are joined together in a hierarchy by using the nested Chinese restaurant process. 
  
  The core assumption of standard topic-model algorithms is that a corpus consisted of $N$ documents. And each document is generated by the processing selecting one topic from $K$ topics with probability $p(topic|doc)$ then selecting one word from $N_w$ distinct words with probability $p(word|topic)$. Then, our problem is translated to estimate $N × K$ probabilities $p(topic|doc)$ and $K × N_w$ probabilities p(word|topic). LDA and PLSI both aim to estimate the values of these probabilities with the highest likelihood of generating the corpus ~\citep{PLSI,LDA,Griffiths06042004,4476690}. Thus, the inference problem is transformed to an optimization problem ~\citep{5563111}. But there exist many competing models with nearly identical likelihoods. Due to the high degeneracy of the likelihood landscape, standard optimization algorithms will more likely infer different models after different optimization runs than infer the model with the highest likelihood,as has been previously reported \cite{5563111,NIPS2009_3854}. A research on the validity of LDA optimization algorithms for inferring topic models also proposed that current implementations of LDA had low validity~\citep{PhysRevX.5.011007}.
 
 Meanwhile, selecting the number of topics $K$ is one of the most problematic modeling choices in finite topic modeling. There is no effective method for choosing $K$ or evaluating the probability of held-out data for various values of $K$ so far. And degree to which LDA is robust to a poor setting of $K$ is not well-understood ~\citep{NIPS2009_3854}. Ideally, if LDA has sufficient topics to model the data set well, an increase in $K$ would not have a impact on the assignments of tokens to topics --i.e., the additional topics should be used with low frequency. For example, if twenty topics is adequacy to exactly model the data, then inferred topic assignments would not be significantly affected by increasing the number of topics to fifty. If this is the case, using large $K$ would not have a improvement on the inference. In another words, we still need a robust $K$. Actually, $K$ could be seem as the rank of the solution space for topic generation. Setting $K$ is same as manually selecting the rank of the solution space, which is obviously not reasonable.
 
 The standard topic-model algorithms focus on the modeling the process of generating documents with topics. In this paper we propose an approach to get an initial guess of topics from the distribution of words and documents. If we think about an easy problem, in which one word can only belongs to one topic. Generating topics from corpus closely approximates to the processing of community detection in networks. A substantial amount of work in the area of community detection in networks has proposed effective algorithms to reveal the struct of the network only using the original information of the network without other prior knowledge. So we create a network of words in the corpus and detecting the communities of the network as the initial guess for topics, then refine these coarse topics. And the words with top $p(word|topic)$ in the topics extracted by HLSM are interesting, it seems that HLSM distinguishes the topics in a more concrete level. For example, image, jpeg, gif" and ''3d, graphic, ray" will be assigned to different topics.

  The contribution of this paper can be summarized as follows:
  \begin{itemize}
  \item Propose a novel approach to constructing network of words closely related to the latent topic space.
  \item Adapt approaches from community detection in networks to initial hierarchical topic generation, and also propose a method to further refine the topics.
  \item To evaluate the effectiveness of the proposed approach, we conducted experiments on several real-world text data sets. The experimental results demonstrate that our approach provides greatly improvements in terms of documents classification. 
  \end{itemize} 
  
 \section{Hierarchical Latent Semantic Mapping}
Hierarchical Latent Semantic Mapping (HLSM) is a network approach to topic modeling. Similar to the well-known topic models, each document is represented as a mixture over latent topics. The key feature that distinguishes the HLSM model from the existing topic models is that HLSM directly clusters words and defines each cluster as a topic, then refines these initial topics, thus HLSM estimates the probability distributions $p(word|topic)$ in a novel process.
 
  The HLSM model infers topics as the following steps:
   \begin{figure}[!h]
\centering
  \includegraphics[width=0.58\linewidth]{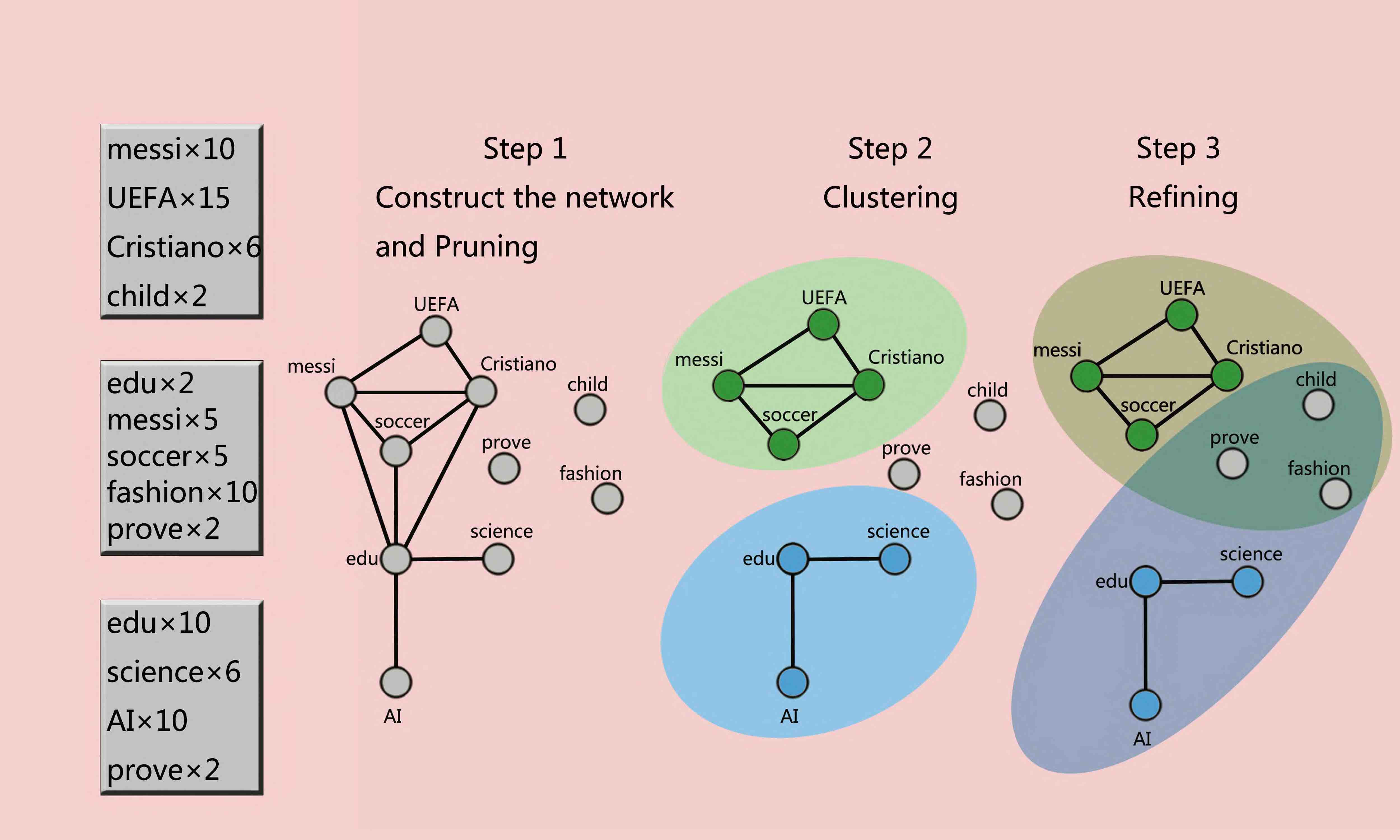}\\
  \caption{Illustration of the HLSM algorithm.}\label{fig:hlsm}
\end{figure}
\begin{enumerate}[step 1.]
\item \emph{Construct the unipartite network}.we calculate the association between each pair of words that co-occur in at least one document. Then we construct the unipartite network in which words are connected with the association above the threshold.
\item \emph{Clustering of words hierarchically}.The words in the unipartite network are connected by the association in the latent topic space. Naturally we suppose that topics in the corpus will give rise to communities of words in the network. Thus we use the \emph{Hierarchical Map Equation} \citet{10.1371/journal.pone.0018209} to detect the communities. And in most of corpus, topics come in the form of multiple levels of abstraction. Abstract topic consists of several concrete topics. Thus we detect some massive communities corresponding to the abstract topics, then we detect minor communities, which correspond to the concrete topics, from the massive communities. We take the communities as a prior guess for the number of topics and word composition of each of the topics used to generate the documents.\emph{It is worth noting} that we do not set the number of levels and the number of communities for each level. Hierarchical Map Equation can reveal the multilevel organization in the network of words automatically.
\item \emph{Refine the prior guess}. After the \emph{last} level of clustering of words, one may get some single communities of words, and in the step 2, one may get some single words not in the network. Thus the prior topics detected in step 2 are rough, we refine the topics using a PLSA-like likelihood optimization.
\end{enumerate}

\subsection{Construct the unipartite network}

  The association between words must be closely related to the topics to ensure the validity of clustering words based on this network. But the topics are latent, and all observations are the words collected into documents. If we assigns topics to documents artificially with prior human knowledge, one can observe that documents share the same topics also are more likely to share some words. Naturally we can believe that the words co-occur in many documents share the same topic, in another word these words are more similar in the latent topic space. 
  To calculate the association between words in the latent topic space. Like the core idea of Latent Semantic Analysis (LSI), we map words to a vector space of reduced dimensionality based on a \emph{Singular Value Decomposition} (SVD) of the co-occurrence matrix $M$, which \emph{each row $i$ corresponds to a word, each column $j$ to a document} in which the word appeared, and each matrix entry $M_{ij}$ corresponds to the number of occurrences of word $i$ in document $j$. 
  
  Starting with the standard SVD given by  
  \begin{equation}
  M = U \Sigma V^t,
   \end{equation} 
   the diagonal matrix $\Sigma$ contains the singular values of M. The approximation of $M$ is computed by setting all but the largest $K$ singular values in $\Sigma$ to zero (= $\widetilde{\Sigma}$), which is rank $K$ optimal in the sense of the $L_2$-matrix norm.
   
One obtains the approximation
    \begin{equation}
    \widetilde{M} = U \widetilde{\Sigma} V^t  \approx  U \Sigma V^t = M,
   \end{equation} 
   
   The corresponding low-dimensional latent vectors will typically not be sparse, while the original high-dimensional Matrix $M$ is sparse. This implies that one can calculate meaningful association values between pairs of words in the latent topic space. In HLSM, we calculate the cosine similarity between the rows of $U\widetilde{\Sigma}$ as the association of each pair of words in the latent topic space, and connects word $i$ and $j$ with this association $S(i,j)$ :
    \begin{equation}
    W = U\widetilde{\Sigma},\\
    \ \nonumber\\
   S(i,j) = \frac{\langle W_i \cdot W_j \rangle }{ \| W_i \| \cdot \| W_j \|}.
   \end{equation}
  
  After calculating all the values of connections. Suppose that the association values between some pair of words are so low that we presume these connections are noise. One can set a threshold of $q$ to purne the connections lower than $q$.
  
\subsection{Clustering words hierarchically} 

  In most of corpus, the structure of topics is not simple and always can be multiple levels. Some concrete topics sit under a same abstract topic. For example, words in a corpus focusing on ``soccer" might be drawn from the topics ``stars", ``matches", ``history of soccer", etc. 
  
  We construct the network of words based on the association between words in the latent topic space. If the original structure of topics is multiple levels, the network should also have a multilevel structure. To reveal communities at multiple levels, we choose the \emph{Hierarchical Map Equation} \citet{10.1371/journal.pone.0018209}. It is worth noting that we do not set the number of levels and the number of communities for each level. Instead Hierarchical Map Equation can reveal the multilevel organization in the network of words automatically.
  
  The Map Equation proposed the duality between finding community structure in networks and minimizing the specification length of a random walker's movements on a network. For a given network partition, the map equation definiens the limit $L(M)$ of how laconic one can describe the trajectory of this random walk in theory.

 The core idea of map equation is that if the random walker tends to stay in some blocks of the network for a long time, the code used for specification can be compressed. Therefore, when the proxy for real flow random walk in the network, estimating the minimum map equation over all possible network partitions could reveals the structure of the network with respect to the dynamics on the network.
 
  In our problem, for a hierarchical network $M$ of $n$ nodes, each node corresponds to one word, segmentated into $m$ modules. There is a a submap $M^i$ with $m^i$ submodules in one modules. Correspondingly, there is a submap $M^{ij}$ with $m^{ij}$ submodules in each each submodule $ij$, and so on.
  
  The corresponding hierarchical map equation is
 \begin{equation}
  L(M) = q_{switch}H(Q) + \sum_{i=1}^mL(M^i)
  \end{equation}
  with the specification length of submap $M^i$ at intermediary levels given by
   \begin{equation}
  L(M^i) = {q^i}_{switch}H(Q^i) + \sum_{j=1}^{m^i} L(M^{ij})
  \end{equation}
  and at the final modular level by
  \begin{equation}
  L(M^{ij...k}) = {p^{ij...k}}_{in} H(P^{ij...k})
  \end{equation}
  
 Weight of codebook depends on the rate of use of it, and $L(M)$ is the sum of average length of codewords for each codebook. $H(Q)$ is the average length of codewords in the index codebook according to the rate of use of it, while the entropy terms depends on the rate at which the codebooks are used. On any given step the random walker switches the \emph{first} level modules at probability of $q_{switch}$, while $q_{switch}$ is the rate of index codebook is used.
  
\begin{figure}[!h]
\centering
  \includegraphics[width=0.4\textwidth]{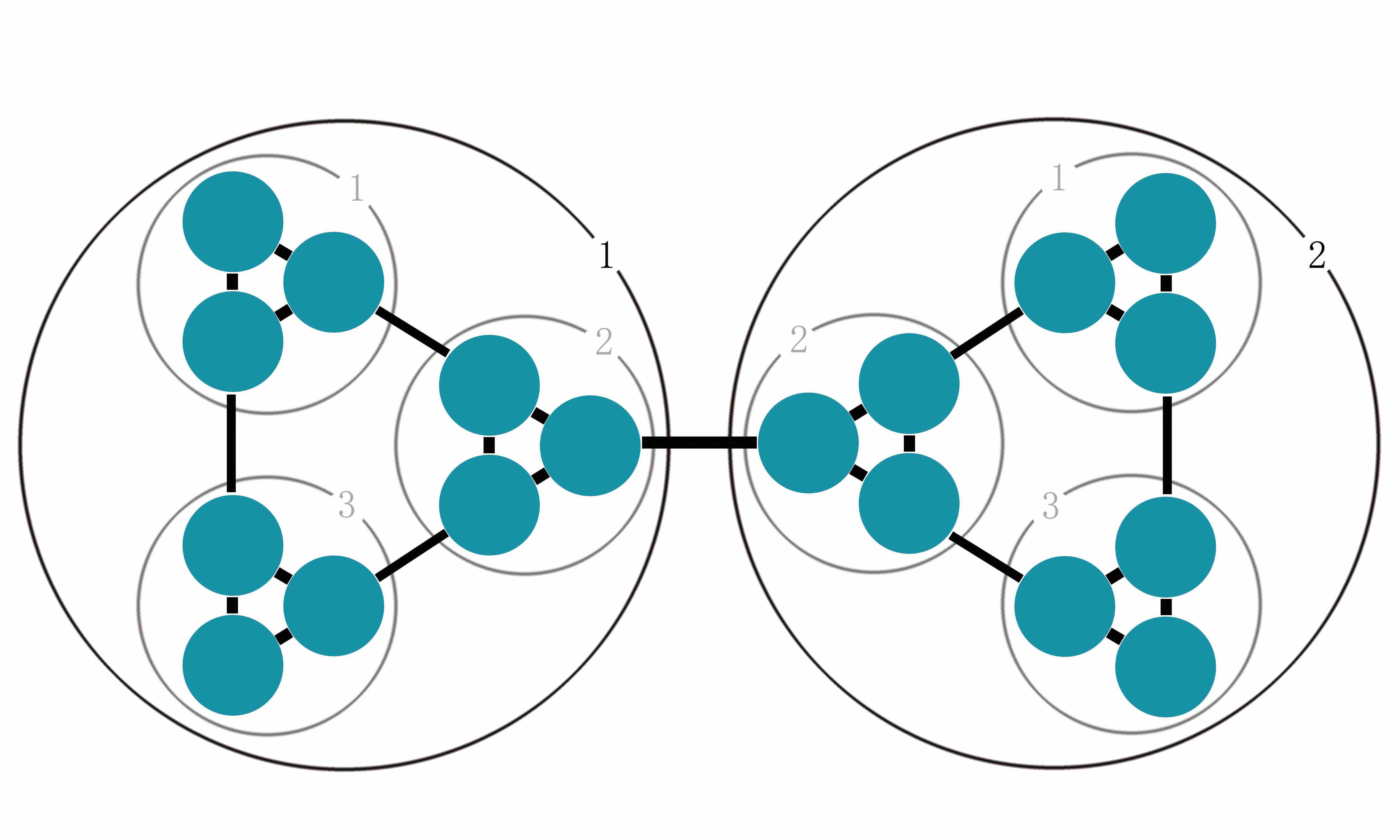}\\
  \caption{Example for Minimizing the map equation over all network partitions gives an optimal clustering of the network with respect to the dynamics on the network.}
\label{fig:mapequation}
\end{figure}

  At each submodule level, $H(Q^i)$ is the average length of the codewords according to the using rate in the subindex codebook and $q_{switch}^i$ is the rate of codeword use for entering the $m_i$ submodules or exiting to a higher level. At the last level, $H(P^{ij...k})$ is the average length of the codewords according to the using rate in the submodule codebook and $p^{ij...k}_{in}$ is the rate of codeword use for visiting nodes in submodules $ij . . . k$ or exiting to other submodules. The problem of seeking the hierarchical structure that best represents the structure is translated to finding the hierarchical partition of the network with the minimum map equation.  Fig.\ref{fig:mapequation} illustrates an example for map equation.
 
  In this example we can assume that all weights for connections in the network are equal, thus all rates can be calculated by counting
links and normalizing. The specification length for an unpartitioned network is $- log_2(1/18) = 4.17 bits$. After the network is partitioned, the codewords of the first level modules are used at a total rate $q_{switch} = \frac{2}{50}$ ( There are 25 lines in the network and 50 possible moves when considering direction, while only 2 moves can switch between the first level module.), while relative rates $Q = \frac{1}{2} , \frac{1}{2}$. And $Q^1 = \frac{2}{8},\frac{2}{8},\frac{3}{8},\frac{1}{8}$, noticing that there is a rate at $\frac{1}{8}$ random walker existing to Module 2, while $q_{switch}^1$ is $\frac{8}{50}$. Thus $L(M)$ is:\\
\begin{small}
$L(M) = q_{switch}H(Q) + \left\{
\begin{aligned}
q_{switch}^1H(Q^1) & + & \left\{
\begin{aligned}
p_{in}^{11}H(P^{11}) \\
p_{in}^{12}H(P^{12}) \\
p_{in}^{13}H(P^{13}) \\
\end{aligned}
\right.\\
q_{switch}^2H(Q^2) & + & \left\{
\begin{aligned}
p_{in}^{21}H(P^{21}) \\
p_{in}^{22}H(P^{22}) \\
p_{in}^{23}H(P^{23}) \\
\end{aligned}
\right.
\end{aligned}
\right.$
\end{small}\\
L(M) = 0.04 bits + 0.61bits + 2.54 bits = 3.19 bits .
  \subsection{Refine the prior guess}
  Once the network is built, we detect clusters (same as the modules detected by \emph{Hierarchical Map Equation}) of highly associated words using the \emph{Hierarchical Map Equation}. After the \emph{last} level of clustering, we get a hard partition of words, meaning that words can only belong to a single cluster. Actually a word may have multiple senses and multiple types of usage in different context. Consequently if we simply define every cluster as a topic, these rough topics can not provide a reasonable probabilistic interpretation of the corpus in terms of the latent topic space. Therefore we propose a method to further refine these rough topics. 
  
  We now discuss how we can compute the distributions $p(topic|doc)$ and $p(word|topic)$, given a partition of words. In the prior partition of words, we define every cluster as a topic. In fact, each word in the network can sit in only one module after the \emph{Hierarchical Map Equation} processing. Therefore, $p(t|w) = \delta_{t,w}$ . $\delta_{t,w} = 1$ only if the word $w$ sits in the module, which corresponds to the topic $t$ . For other topics $\bar{t}$ , $\delta_{\bar{t},w} = 0$ . Noticing that in this step word $w$ can only belongs to one topic t, so $p(w,t)=p(w)$ , thus:
   \begin{equation}
   p(w|t) = \frac{p(w)} {\sum_w p(w)\times \delta_{t,w}}  \mbox{    and    }  p(t|d) = \frac{1}{L_d} \sum_{w} w_w^d \delta_{t,w} .
  \end{equation}
$L_d$ is the number of words in document $d$, $w_w^d$ is the number of times word $w$ occurs in the document $d$. It is also useful to introduce $n(w,t) = L_C \times p(w,t)$, which is the number of times topic $t$ was chosen and
word $w$ was drawn.$L_C$ is the number of the words in the corpus.
 So far, the PLSA-like likelihood of our model is:
 \begin{equation}
 \begin{split}
 L = log (\prod_{w,d}p(w,d)) = log (\prod_{w,d} \sum_{t}p(w|t)p(t|d))  \\= \sum_d \sum_w w_w^d \times log(\sum_{t}p(w|t)p(t|d)) \ .
 \end{split}
 \end{equation}
  We can improve this likelihood by simply making documents more specific to fewer topics. For that our optimization algorithm simply finds, for each document, words assigned with some infrequent topics and reassigns the most significant topic in that document to these words.
  \begin{enumerate}[1.]
  \item For each document $d$, we find the most \emph{significant} topic $t_s$ with the smallest $p$-value, considering a null model where each word is independently sampled from topic t with probability $p(t) = \sum_{w}p(w)p(t|w)$. Calling $x$ the number of words which actually come from topic $t$, ($x = L_d \times p(t|d)$ , see Eq . (6) ) , the $p$-value of topic t is then computed using a binomial distribution, $B(x; Ld, p(t))$. Obviously $p$-value represents the significance of the word better than $x$, which only depends on the $p(t|d)$.
  \item For each document $d$, recall that after the step 2 we may get some single words not in the network. We simply assign these words to the most significant topic $t_s$ and we can calculate a baseline of the PLSA-like likelihood L(see Eq .(7)).
  \item For each document $d$, we define the \emph{infrequent} topics $t_{in}$ simply as those which occur with probability smaller than a parameter: $p(t_{in}|d) < \eta$. We assign the most significant topic $t_s$ to the words which belong to the all infrequent topics $t_in$. The $p(t_s|d)$ will be incremented by the sum of all $p(t_{in}|d)$, while all $p(t_{in}|d)$ are set to zero. Similarly, $n(w,t_{in})$(see above) will be decreased by $w_w^d$ for each word w which belongs to an infrequent topic, and $n(w,t_s)$ is increased accordingly.
  \item After previous step for all document, we compute:
  \begin{equation}
 \begin{split}
 p(w|t) = \frac{n(w,t)}{\sum_w n(w,t)}
  \end{split}
 \end{equation} and the likelihood of model, $L_{\eta}$, where we made explicit its dependency on $\eta$. We pick the model with maximum $L_{\eta}$ by looping over all possible values of $\eta$ (from 0\% to 50\% with steps of 1\%).
  \end{enumerate}
  
  HLSM estimates the probabilities $p(w|t)$ and $p(t) = \sum_{w}p(w)p(t|w)$ from training data set, and calculates $p(t|w) = \frac{p(t)p(w|t)}{p(w)}$, for a new document from held out data set, $p(w|t)$ won't be changed, $p(t|d)$ can be calculated by :
   \begin{equation}
 \begin{split}
 p(t|d) = \frac{\sum_wp(t|w)}{L_d}
 \end{split}
 \end{equation}
 
  HLSM fixed the probabilities $p(w|t)$ and $p(t)$ after the training process, and hence is plagued by overfitting. It will be a shortcoming of the HLSM model, when the scale of the training data set is small.
  
  \section{Experimental Evaluations}
  HLSM is a topic model towards collections of text corpora. It can be applied to lots of applications such as classifying, clustering, filtering, information retrieval and related areas. Follow Blei's idea \citet{LDA}, in this section, we investigate two important applications: document modeling and document classification.
  \subsection{Document Modeling}
  The goal of document modeling is to generalize the trained model from the training dataset to a new dataset. The documents in the corpora are unlabeled, our goal is density estimation, thus we wish to obtain high likelihood on a held-out test set. In particular, we computed the \emph{perplexity} of a held-out test set to evaluate the models. Models which yield a lower perplexity are considered to achieve a better generalization performance because the model is less surprised by a portion of the datasets which the model have never seen before. Formally, for a test set of $M$ documents, the perplexity is defined as:
 \begin{equation}
  perplexity(D_{test}) = exp \left \{ \frac{-\sum_{i=1}^Mlogp(d_i)}{\sum_{i=1}^ML_i}\right \}
  \end{equation}
   \begin{figure}[h]
\centering
  \includegraphics[width=0.5\textwidth]{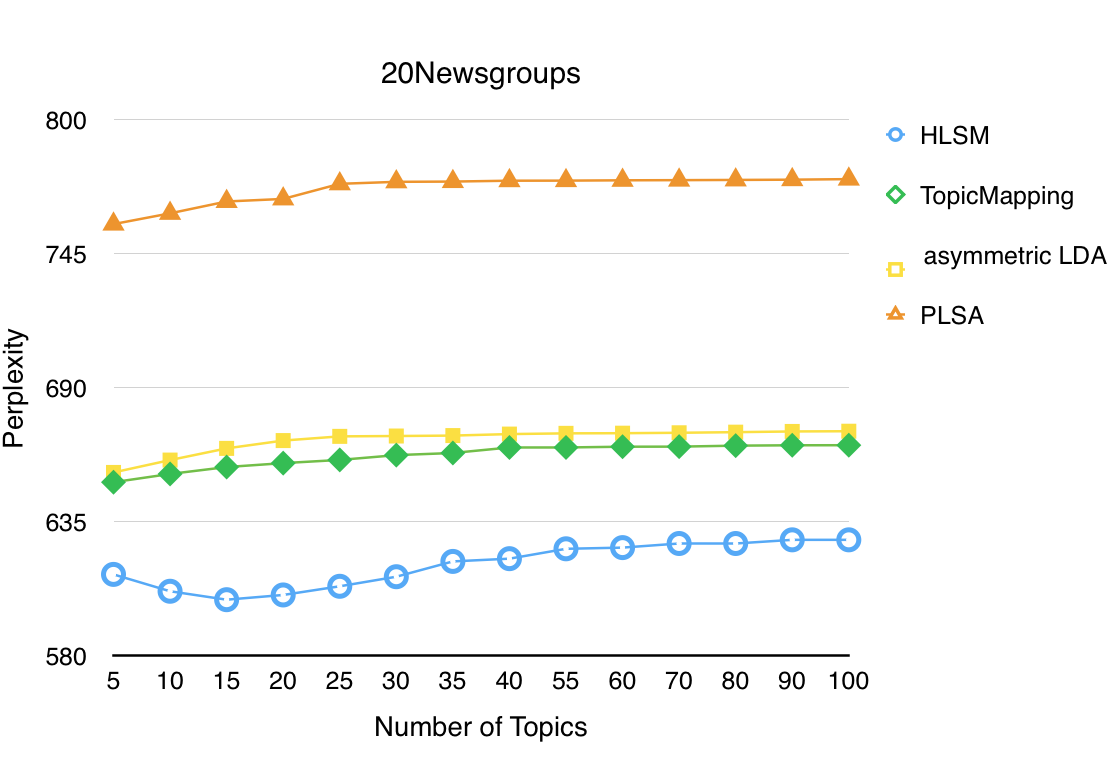}\\
  \caption{Perplexity comparisons on the 20Newsgroups dataset.}\label{fig:perplexity}
\end{figure} 
    We conduct this experiment on a subset of the 20Newsgroups data set, which has been widely used for evaluating the performance of cross-domain text classification algorithms. It contains nearly 20,000 newsgroup documents which have been evenly partitioned into 20 different newsgroups. We chose 3878 documents (we filtered some little documents) from domain comp.graphics, com.sys.mac.hardware, sci.crypt, and sci.med as our dataset used in the evaluation. We held out 20\% of the corpus for test purpose and trained the models on the remaining 80\%. In data preprocessing, we removed 163 stop words in standard list and the words occurrences less than 3 times from each corpus. We compare HLSM against PLSA, asymmetric LDA and TopicMapping. The initial $\alpha$ for asymmetric LDA was set to 0.01 for all topics.
    
  Fig. \ref{fig:perplexity} shows the perplexity results where the number of the topics varies from 5 to 100. As can be seen, the HLSM model achieves slight improvement in terms of perplexity, while TopicMapping is close to asymmetric LDA. Experiment shows that the prior guess of HLSM makes great difference on the topic generation. 
  Table \ref{table:topics} presents the examples of top 12 extracted topics on data set \emph{Comp and Sci}, some topics with lower probability were not exhibited. We sorted the words with the learned topic-word probability. By examining the topical words, we can observe that the words in the same topic are always semantically relevant. For example, Topic 1 is about Mac hardware, and one domain in the data set Comp and Sci is comp.sys.mac.hardware, respectively. It is noteworthy that, some topics look similar in abstract level, but there are still some distinctions between them. For instance, words in Topic 2 and Topic 4 are semantically relevant but Topic 2 is more related to medical treatment, while Topic 4 probably describes some reports about disease. The result shows that our method can effectively identify the correlations between domain-specific features from different domains. Furthermore, our method can extracted narrow topics under the level of domain. And we conduct the next experiment on the whole 20Newsgroups data set.
    
\begin{table}[t]
\caption{Examples of Top 12 Topics Extracted by HLSM on Data Set Comp and Sci}
\label{table:topics}
\begin{center}
\begin{tabular}{llllll}
topic: 1  & topic: 2 & topic: 3 &  topic: 4 &  topic: 5 & topic: 6 \\
$p(t)$ : 0.0801 & $p(t)$ : 0.0672& $p(t)$ : 0.0662 & $p(t)$ : 0.0619 & $p(t)$ : 0.0607 & $p(t)$ : 0.0600\\
\hline
mac & doctor & clipper & medic & food & imag\\
doe & patient & phone & health & msg & jpeg\\
system & vitamin & chip & 1993 & diet & file\\
speed & medic & encrypt & diseas & eat & format\\
price & candida & govern & hiv & weight & gif\\
hardware & treatment & onli & report & effect & program\\
\hline
topic: 7  & topic: 8 & topic: 9 &  topic: 10 &  topic: 11 & topic: 12 \\
$p(t)$ : 0.0562 & $p(t)$ : 0.0557 & $p(t)$ : 0.0507 & $p(t)$ : 0.0463 & $p(t)$ : 0.0455 & $p(t)$ : 0.0441\\
\hline
imag & drive        & key        & anonym & nsa     & 3d\\
data & disk          & encrypt & email & writes       & graphic\\
system & system & messag & internet & govern & file\\
packag & work     & secur    & post & articl         & object\\
sourc & scsi         & pgp       & comput & david     & ray\\
code & machin   & attack    & inform & trust       & model\\
\hline
\end{tabular}
\end{center}
\end{table}

  
\begin{table}[t]
\caption{The Test Classification Accuracy on The Data Sets Generated from 20Newsgroups}
\label{table:classification}
\begin{center}
\begin{tabular}{llllll}
\multicolumn{1}{c}{\bf Data set}  &\multicolumn{1}{c}{\bf PLSA}&\multicolumn{1}{c}{\bf LDA}&\multicolumn{1}{c}{\bf asymmetric LDA}&\multicolumn{1}{c}{\bf TopicMapping}&\multicolumn{1}{c}{\bf HLSM}
\\ \hline \\
Data set & PLSA & LDA & asymmetric LDA & TopicMapping & HLSM\\
 Comp and Sci & 0.761 & 0.771 & 0.792 & 0.831 & \textbf{0.855}\\
Comp and Talk & 0.785 & 0.790 & 0.813 & 0.846 & \textbf{0.871}\\
Comp and Rec & 0.770 & 0.776 & 0.781 & 0.834 & \textbf{0.853}\\
Sci and Rec & 0.724 & 0.723 & 0.767 & 0.803 & \textbf{0.822}\\
Talk and Rec & 0.811 & 0.802 & 0.832 & 0.821 & \textbf{0.876}\\
Talk and Sci & 0.804 & 0.811 & 0.839 & 0.847 & \textbf{0.867}\\
Average & 0.766 & 0.779 & 0.804 & 0.834 & \textbf{0.857} \\
\end{tabular}
\end{center}
\end{table}
    \begin{table}[t]
\caption{Data Sets Generated from 20Newsgroups }
\label{table:dataset}
\begin{center}
\begin{tabular}{ll}
\multicolumn{1}{c}{\bf Data set}  &\multicolumn{1}{c}{\bf Domain}
\\ \hline \\
 Comp and Sci & comp.graphics, comp.sys.mac.hardware, sci.crypt, sci.med\\
Comp and Talk & comp.os.ms-windows.misc, comp.sys.ibm.pc.hardware, talk.politics.mideast, talk.politics.misc\\
Comp and Rec & comp.graphics, comp.sys.ibm.pc.hardware, rec.motorcycles, rec.sport.baseball\\
Sci and Rec & sci.crypt, sci.med, rec.autos, rec.sport.baseball\\
Talk and Rec & talk.politics.mideast, talk.politics.misc, rec.autos, rec.sport.baseball\\
Talk and Sci & talk.politics.misc, talk.religion.misc, sci.crypt, sci.med\\
\end{tabular}
\end{center}
\end{table}

 \subsection{Document classification}
In the text classification problem, topic models are wished to classify a document into two or more mutually exclusive classes. The choice of features is a challenging aspect of the document classification problem. By representing the documents in terms of latent topic space, the topic models can generate the probabilities $p(t|d)$. If one use the vector of $p(t|d)$ as the feature of documents to fix the text classification problem, the probabilities vector generated by the most effective model can perform better than the probabilities vector generated by other models. 

  To test the effectiveness of HLSM, we compared it with the following representative topic models and chose AC as the evaluation there.
PLSA, symmetric LDA, asymmetric LDA, TopicMapping.
  
   We generated six cross-domain text data sets from 20Newsgroups by utilizing its labeled structure. There are 4 fields in each data set, Table \ref{table:dataset} summarizes the data sets generated from 20Newsgroups. To make the classification problem more effective and convincing, the task was defined as a multi-label classification. 
   
   In these experiments, we estimated the probabilities $p(t|d)$ using the above topic models on all the documents of each data sets, and used the vector of probabilities $p(t|d)$ as the only features to train a support vector machine (SVM) for multi-label classification. For each data set, 20\% of the documents were held out as the test data and we trained a SVM for multi-label classification with the remaining 80\% labeled documents. We used these classifiers to predict the class labels of unlabeled documents in the test data. Notice that there were 4 field in each data set, the classification process was considered as correct only if the document was classified into the original field.

  We did the same data preprocessing as above, and the number of topics in each data set for LDA, PLSA, and asymmetric LDA was set to 4. Table \ref{table:classification} summarizes the classification performance on each data set, the first three row shows the best accuracy while the number of topics for LDA, PLSA, and asymmetric LDA varies. The last row of the table shows the average accuracy over all data sets. From the table we can observe that HLSM outperformed all other topic models on six data sets. 
 
\section{Conclusion}
A topic model HLSM is presented in this paper to apply an approach from the area of community detection to topic generation. We apply the HLSM model to several document collections for document modeling and document clustering, and the experimental comparisons against state-of- the-art approaches demonstrate the promising performance. In particular, in the area of community detection, a substantial amount of work has been done on stochastic block models, which tries to fit a model to reveal community structure in networks. We believe this work, which is similar to topic model in spirit, would offer new insights into topic modeling.

\bibliography{iclr2016_conference}
\bibliographystyle{iclr2016_conference}
\end{document}